\documentclass[11pt,a4paper]{article}
\usepackage[hyperref]{eacl2021}
\usepackage{times}
\usepackage{latexsym}
\usepackage{changepage}
\usepackage[inline]{enumitem}

\usepackage{microtype}
\usepackage{graphicx}
\aclfinalcopy 

\title{Federated Word2Vec: Leveraging Federated Learning to Encourage Collaborative Representation Learning
\thanks{\ \ This project has received funding from the European Union’s Horizon 2020 research and innovation programme under the Marie Skłodowska-Curie grant agreement No 813162. The content of this paper reflects the views only of their author (s). The European Commission/ Research Executive Agency are not responsible for any use that may be made of the information it contains.}}
  
\author{Daniel Garcia Bernal\footnotemark\ \ \ Lodovico Giaretta\footnotemark[2]\ \ \  \v{S}ar\={u}nas Girdzijauskas\footnotemark[2]\ \ \ Magnus Sahlgren\footnotemark[3]\\
\footnotemark[2]\ \ KTH Royal Institute of Technology \\
\footnotemark[3]\ \ RISE Research Institutes of Sweden \\
{\tt \{danigb,lodovico,sarunasg\}@kth.se \hspace{1cm} magnus.sahlgren@ri.se}
}

\date{}

\begin{document}
\maketitle
\begin{abstract}
Large scale contextual representation models have significantly advanced NLP in recent years, understanding the semantics of text to a degree never seen before. However, they need to process large amounts of data to achieve high-quality results. Joining and accessing all these data from multiple sources can be extremely challenging due to privacy and regulatory reasons. Federated Learning can solve these limitations by training models in a distributed fashion, taking advantage of the hardware of the devices that generate the data. We show the viability of training NLP models, specifically Word2Vec, with the Federated Learning protocol. In particular, we focus on a scenario in which a small number of organizations each hold a relatively large corpus. The results show that neither the quality of the results nor the convergence time in Federated Word2Vec deteriorates as compared to centralised Word2Vec.
\end{abstract}

\section{Introduction}

A central task in NLP is the generation of word embeddings to encode the meaning of words and their relationships in a vector space. This task is usually performed by a self-supervised Machine Learning (ML) model such as Word2Vec~\cite{w2v}, ELMo~\cite{elmo} or BERT~\cite{bert}, with access to a large corpus of documents as input. These representations can then be used to perform advanced analytics on textual data. The larger and more complete the corpus is, the more accurate the representations will be.

As such, it can be useful for multiple organizations to collaborate with each other, each providing access to their corpora, in order to obtain the best results. However, different organizations typically cannot easily share their data, as they have to protect the privacy of their users and the details of their internal operations, or might be bound by external laws preventing the sharing of the data. One way organizations could overcome these issues is by employing Federated Learning protocol~\cite{fl} to generate a global model without sharing any data. It is therefore fundamental to assess the performance, quality and privacy preservation characteristics of such approach.

\subsection{Distributed vs data-private massively-distributed approaches}

Datacenter-scale ML algorithms work with vast amounts of data allowing the training of large models, exploiting multiple machines and multiple GPUs per machine. Currently, GPUs offer enough power to satisfy the needs of state of the art models. However, this exposes another important aspect for consideration - Data Privacy. In recent years, users and governments have started to be aware of this issue, publishing new and stricter regulations, such as the GDPR~\cite{EUdataregulations2018}. Companies started making efforts to shield themselves from any security leak that could happen in a centralised system. This created a need to move the research towards distributed architectures where the data is not gathered by a central entity. 

The fast development of smart devices, the growth of their computational power and of fast Internet connections, like 4G and 5G, enable new approaches that exploit them to train distributed models. This solution is currently not at the same scale of resources that a datacenter can offer yet, but the research and development of edge devices is making it feasible.

For these reasons, researchers are exploring the possibilities of different massively-distributed training designs. These new designs should offer scalability, ensure data privacy and reduce large traffic of data over the network. The main massively-distributed approach to large-scale training that fulfills all  these requirements is Federated Learning \cite{fl}.

\subsection{Federated Learning in a small collaborative NLP scenario}

Federated learning is frequently applied in a commercial, global-scale setting as an on-device learning framework. In this common scenario, many small devices, like smartphones, collaborate and grant access to their data, typically small due to storage limitations, to train a higher-quality model. These models are usually NLP applications to make suggestions \cite{fl_1} or predict a user behaviour \cite{fl_2}. 

However, Federated Learning could be applied in different contexts. This paper shifts the focus to address a new scenario where the users are a small number of large organizations with access to large corpora, which cannot be shared or centralised. These organizations are willing to cooperate in order to have access to a larger corpus with diverse topics and to overcome the very strict data privacy policies. A practical example could be a group of government agencies, each of which has only access to sensitive documents in a specific domain (e.g. taxes), which alone would not be sufficient for high-quality training.

\section{Federated Word2Vec}

Federated Learning addresses the privacy concerns as the data is not shared between the organizations. It stays in the node of the owner and the information transferred through the network is only the gradient of the model. It also avoids the expensive transfer of the training data, replacing it with the repeated transfer of gradients, the total size of which is, generally, less than that of the dataset. And even if repeated gradient exchanges were to surpass the size of the dataset, their transfer would be spread on a long time and divided in smaller batches, so it would not delay the training as much as having to send a huge dataset before starting. The only common point that all the nodes share in this architecture is the existence of a central node which oversees the training process, directing the data transfers and merging the contributions of each node. Having the central node can facilitate the inclusion of additional safety measures to shield the training process from malicious attacks \cite{secure_fl}. 

In Federated Word2Vec, each organization owns a private dataset, which means that words that appear in the corpus of one organization may not be present in the one of another. This an issue as the input vocabulary must be common to all local models, so that the gradients can be aggregated. Preserving the privacy of the content of the text is very important, so this paper will overcome the aforementioned issue through a strategy that consists of a common agreement of all the participants in a global vocabulary. All must agree on a fixed vocabulary size $N$ and a minimum threshold of occurrences $T$. Each participant must provide a list of their top $N$ words that appear in their respective texts surpassing $T$ occurrences. The privacy is preserved as the organizations only share a set of isolated, unsorted words. However, the question arises of how to merge these sets. There are two operations that can be applied to produce the final vocabulary: intersection and union. We use, and recommend to use, the union operation. The final vocabulary is larger than the initial size $N$, but all organizations keep their relevant words. Although this approach requires more time to converge due to many words appearing only in certain datasets, the words meaning and knowledge return to the participants is enriched.

Once the participants receive the common agreed vocabulary, the training process can start following the FederatedSGD algorithm from \cite{fl}. The gradient is transferred from all the external nodes to the main node in each iteration. The average gradient is calculated and transferred back to perform the updating process.

\section{Experiments}

\subsection{Data collection}

We generated topic-related datasets collected from Wikipedia articles and organised in different sizes. Two Wikipedia dumps were downloaded to satisfy the aforementioned datasets characteristics: a partial dump with a compressed size of 180 MB to simulate small organizations with short corpora; and a 16 GB compressed file with all the text content published in Wikipedia. From the whole Wikipedia dump, the Wikipedia Extractor script \cite{extractor} is used with a tag to filter categories and prepare 5 different datasets divided by topic. The chosen topics are: biology, history, finance, geography, and sports. Although the themes are quite specific, some articles can appear in more than one dataset because of the distribution of the Wikipedia tree of categories. So, if an article is tagged with the biology category, it is included in the dataset of biological content. In order to simulate a larger number of organizations, every topic is split between two organizations.

\subsection{Setup}

The simulation is performed sequentially on a single machine and with a single GPU, a Nvidia Quadro RTX 5000. This limits the possibility to study the influence of the network, that is thus not covered in this work. The hyperparameters were set to sensible values based on existing literature and should provide a good compromise of training speed and quality. We use a \textbf{fixed batch size} of 2,048 samples. It is important to notice that one iteration of Federated Word2Vec processes as much data as $N$ iterations of centralized Word2Vec, because each of the $N$ nodes processes one batch in parallel during each iteration, in our study $N$ is equal to 10 nodes. The \textbf{embeddings size} is fixed to 200. The number of \textbf{negative samples} per batch is 64, a small amount of negative samples, compared to the total batch size, but sufficient to achieve good results \cite{w2v}. The \textbf{vocabulary size} is 200,000 unique words, with a minimum threshold of 10 occurrences.

\section{Results}

\subsection{Proving convergence of the model with small datasets}

\begin{figure}[!ht]
  \begin{center}
    \includegraphics[width=1\linewidth]{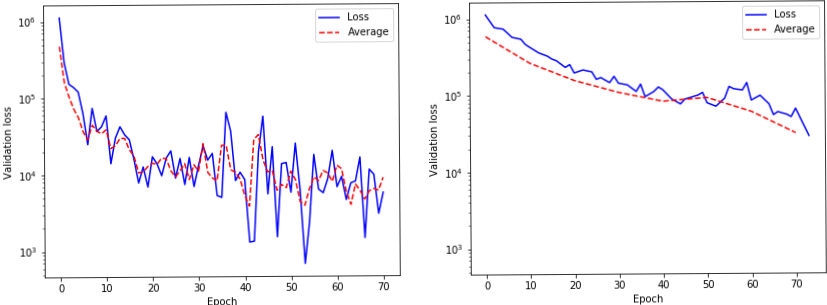}
  \end{center}
  \caption{On the left, validation loss per epoch in a full execution of Word2Vec. On the right, validation loss per epoch of the first 500.000 iterations of Federated Word2Vec. The red lanes represent the average of the validation loss calculated by aggregating all previous values from each epoch. Y-axis is in logarithmic scale.}
  \label{fig:val_base}
\end{figure}

Figure \ref{fig:val_base} compares the validation loss of Federated Word2Vec with that of a baseline, centralized implementation. In order to compare the two models when both have processed the same amount of data, Federated Word2Vec is stopped at epoch 70, which is iteration 500,000.

\begin{figure}[!ht]
  \begin{center}
    \includegraphics[width=1\linewidth]{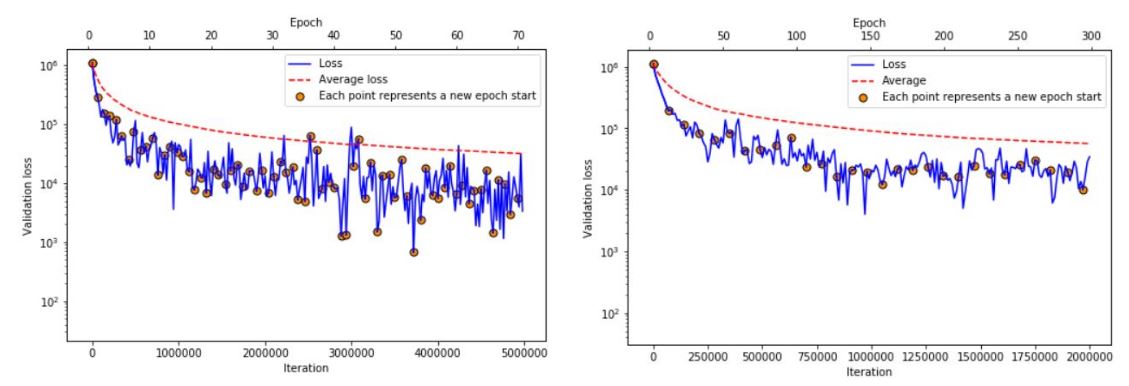}
  \end{center}
  \caption{On the left, validation loss per iteration in a full execution of Word2Vec. On the right, validation loss per iterarions in a full execution of Federated Word2Vec. The red lanes represent the average of the validation loss calculated by aggregating all previous values from each epoch. Y-axis is in logarithmic scale.}
  \label{fig:val_base_1}
\end{figure}

The loss of Word2Vec presents a value of ${\small\sim} 10^{4}$. It is stable, but with a small descending trend. On the other hand, although Federated Word2Vec does not reach the same loss (it is 10 times greater) its trend is clearly decreasing. To check if the trend continues, Figure \ref{fig:val_base_1} illustrates the validation loss in terms of iteration for the full execution, 2 million iterations. The loss keeps going down until 1 million iterations when it stabilises. Overall, the two models provide very similar results. Centralized Word2Vec has a faster initial convergence; however, this might be overcome by adapting the hyperparameters to the distributed setting, for example with learning rate scaling \cite{learning_rate_scaling}.

\subsection{Proving convergence of the model with large datasets}

\begin{figure}[!h]
  \begin{center}
    \includegraphics[scale=0.49]{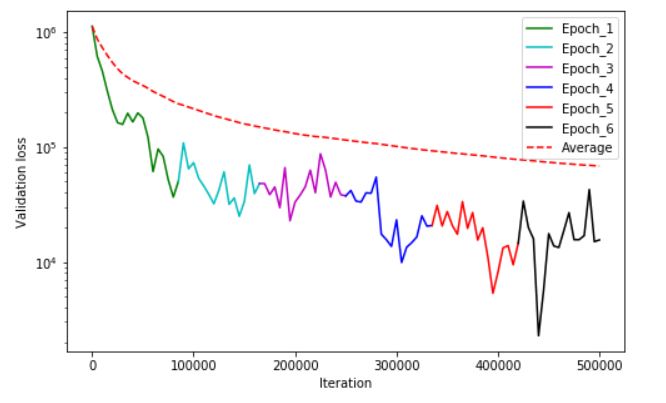}
  \end{center}
  \caption{Validation loss of the first 500.000 iterations of Federated Word2Vec with a larger dataset, divided by epoch. The red lanes represent the average of the validation loss calculated by aggregating all previous values from each epoch. Y-axis is in logarithmic scale.}
  \label{fig:large_loss}
\end{figure}

\begin{figure}[!h]
  \begin{center}
    \includegraphics[scale=0.49]{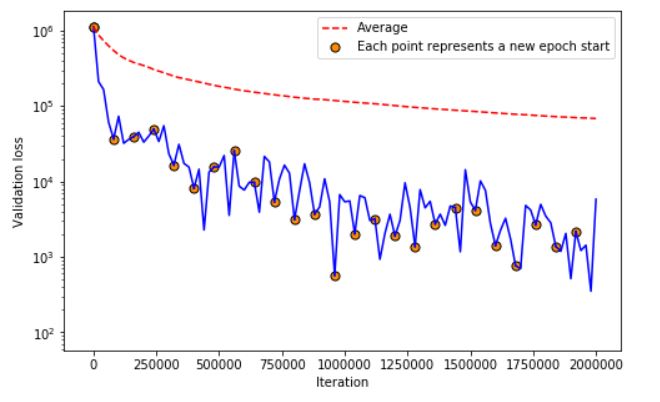}
  \end{center}
  \caption{Validation loss of a full execution of Federated Word2Vec with a larger dataset, represented in blue. The red lanes represent the average of the validation loss calculated by aggregating all previous values from each epoch. Y-axis is in logarithmic scale.}
  \label{fig:last}
\end{figure}

The training of Federated Word2Vec with a large dataset presents improved results compared to the previous graphs. Figure \ref{fig:large_loss} freezes the training in the iteration 500,000 as it was done in Figure \ref{fig:val_base}. The number of epochs is fewer than in the former experiment but the loss presents a clear downward trend with a steeper slope. In Figure \ref{fig:last}, where the execution continues until iteration 2 millions, the loss keeps decreasing reaching values of $10^{3}$, something that did not happen in Figure \ref{fig:val_base_1}.

Consequently, Federated Word2Vec seems to work better with larger datasets as it benefits from learning from multiple sources at the same time. The results show that Federated Word2Vec is not better, and might perform slightly worse, than Word2Vec under the same settings. However, it is proven that Federated Word2Vec has a similar convergence pattern to Word2Vec and easily scales to a large dataset. 

\subsection{How categorised data influence the results}

We then compare collaborative training with Federated Word2Vec to local training by a single organization, which only has access to the \textit{finance} dataset. We analyse the organization of the words in the embedding space, using their cosine distance, by identifying the top-5 closest neighbours for a number of target words, as shown in Table \ref{font-table}.

The most striking finding in this analysis is that clusters are populated with more meaningful words in Federated Word2Vec. This behaviour was expected for the target word \textit{bacteria}, as it does not appear frequently in the \textit{finance} dataset. However, the same situation happens with \textit{market}, presenting meaningless words as the closest neighbours in its community, while the execution of Federated Word2Vec shows more specific context words. 

Moreover, \textit{market} is not an outlier. Most words that are relevant to the \textit{finance} dataset present similar results. The resultant neighbourhood of the word \textit{money} trained with baseline Word2Vec on the financial dataset still presents generic words such as \textit{\{stated, said, there\}}. In contrast, the community generated during the federated training clearly gathers meaning from the finance topic.

These results show the importance of having a full picture of the language to produce high-quality embeddings, even for domain-specific tasks. This, in turn, underscores the need for collaboration among organizations.

\begin{table}
\centering
\small
\begin{tabular}{l l l l l}

\hline \textbf{Word} & \multicolumn{2}{c}{\textbf{W2V}} & \multicolumn{2}{c}{\textbf{Fed W2V}}\\ \hline
\hline

\textbf{Market} & \textbf{Top-5} & \textbf{Dist} & \textbf{Top-5} & \textbf{Dist} \\ \hline
& this & 0.023 & markets & 0.029 \\
& proposed & 0.024 & company & 0.035 \\
& some & 0.024 & share & 0.042 \\
& all & 0.025 & trading & 0.048 \\
& other & 0.025 & assets & 0.049 \\ \hline \hline

\textbf{Bacteria} & \textbf{Top-5} & \textbf{Dist} & \textbf{Top-5} & \textbf{Dist} \\ \hline

& rare & 0.026 & organism & 0.070 \\
& animals & 0.026 & toxic & 0.075 \\
& applied & 0.026 & tissue & 0.077 \\
& result & 0.027 & cells & 0.081 \\
& plants & 0.027 & humans & 0.083  \\ \hline \hline

\textbf{Money} & \textbf{Top-5} & \textbf{Dist} & \textbf{Top-5} & \textbf{Dist} \\ \hline
& stated & 0.028 & paid & 0.045 \\
& said & 0.028 & offer & 0.053 \\
& there & 0.028 & sell & 0.062 \\
& take & 0.029 & cash &  0.071 \\
& help & 0.031 & interest & 0.073 \\

\end{tabular}
\caption{\label{font-table} Top-5 nearest neighbours of each central word, using the cosine distance in the training of W2V with finance dataset and Fed W2V with all 5 datasets.}
\end{table}

\section{Conclusions}

The purpose of this paper was to implement and test the viability of a distributed, efficient, data-private approach that allows a small number of organizations, each owning a large private text corpus, to train global word representations. The results indicate the potential for applicability to real scenarios of collaborative training. The main contributions of this work are 
\begin{enumerate*}
\item the viability of training NLP models like Word2Vec under the Federated Learning protocol with convergence times, at least, at the same level of the widely tested Word2Vec;
\item the importance for organizations to cooperate, as cooperation provides models that are not only globally good, but also locally better than locally-trained models; and
\item the quality of vector representations is not affected by the size of the corpora.
\end{enumerate*}

\noindent

\bibliography{eacl2021}
\bibliographystyle{acl_natbib}

\end{document}